\documentclass[runningheads]{llncs}
\usepackage[T1]{fontenc}
\usepackage{graphicx}
\usepackage{booktabs}
\usepackage{multirow}
\usepackage[export]{adjustbox}
\usepackage{xcolor}
\usepackage{tikz}
\usepackage{amsmath}
\graphicspath{{images/}}

\usepackage{hyperref}
\usepackage{orcidlink}
\renewcommand{\orcidID}[1]{\orcidlink{#1}}

\usepackage{color}

\usepackage{cleveref}

\begin{document}
% Figures for UltraPIPS paper

% Colors used in figures
\definecolor{ClassicalCol}{HTML}{34495e}
\definecolor{ImageNetCol}{HTML}{3498db}
\definecolor{RadiologyCol}{HTML}{e67e22}
\definecolor{UltrasoundCol}{HTML}{2ecc71}

% --- Figure: Spearman Correlations (Fig 1) ---
\newcommand{\figSpearmanCorrelations}{
\begin{figure*}[tb]
    \centering
    \setlength{\tabcolsep}{0.1em}
    \begin{tabular}{@{}ccc@{}}
        & $CAMUS$ & $EchoNet$ \\
        \adjustbox{valign=m}{\rotatebox{90}{(a) $EchoPrime$}} &
        \includegraphics[width=0.5\textwidth, valign=m]{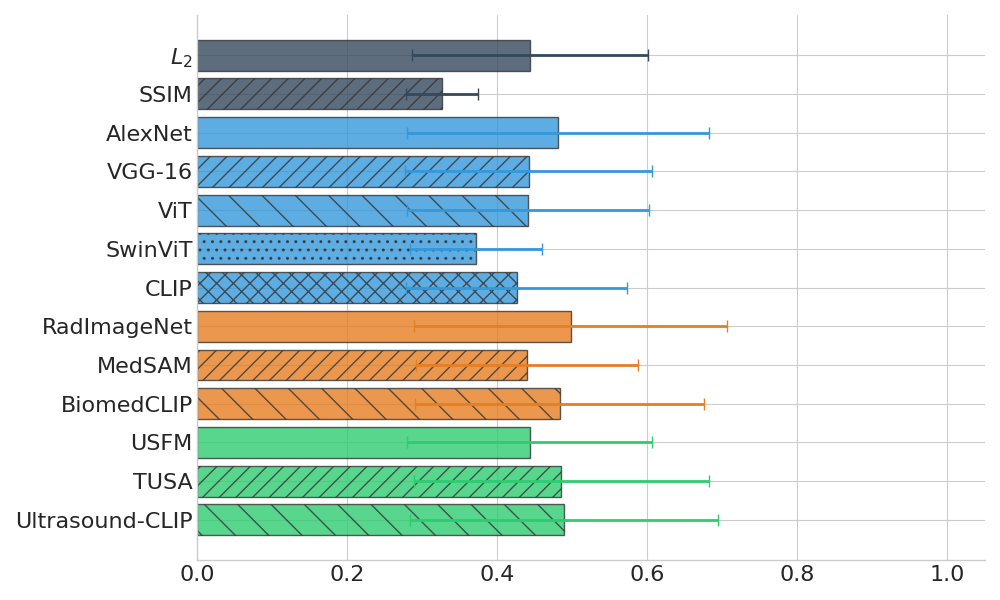} &
        \includegraphics[width=0.5\textwidth, valign=m]{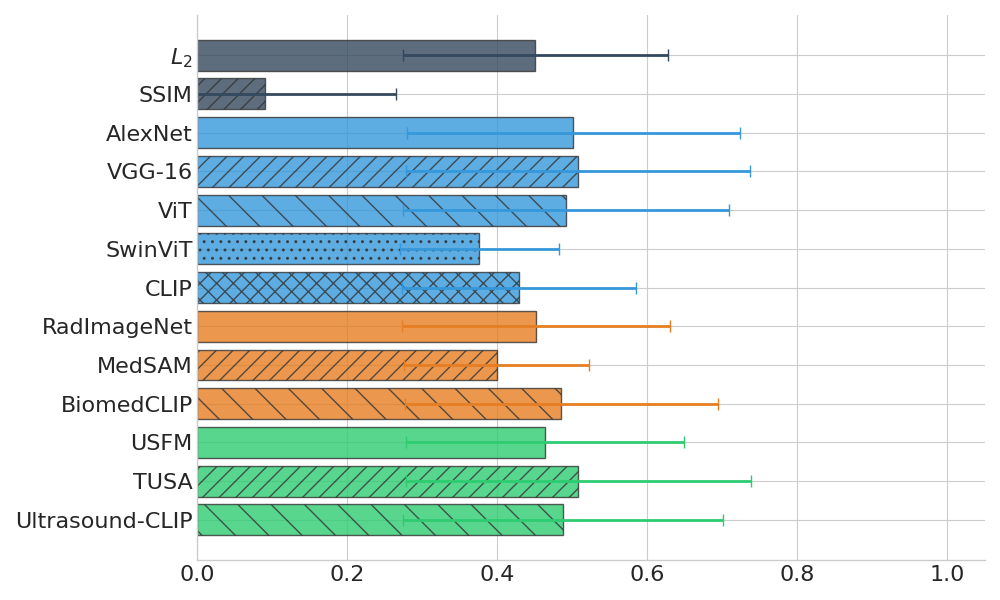} \\
        \adjustbox{valign=m}{\rotatebox{90}{(b) $UltraSam$}} &
        \includegraphics[width=0.5\textwidth, valign=m]{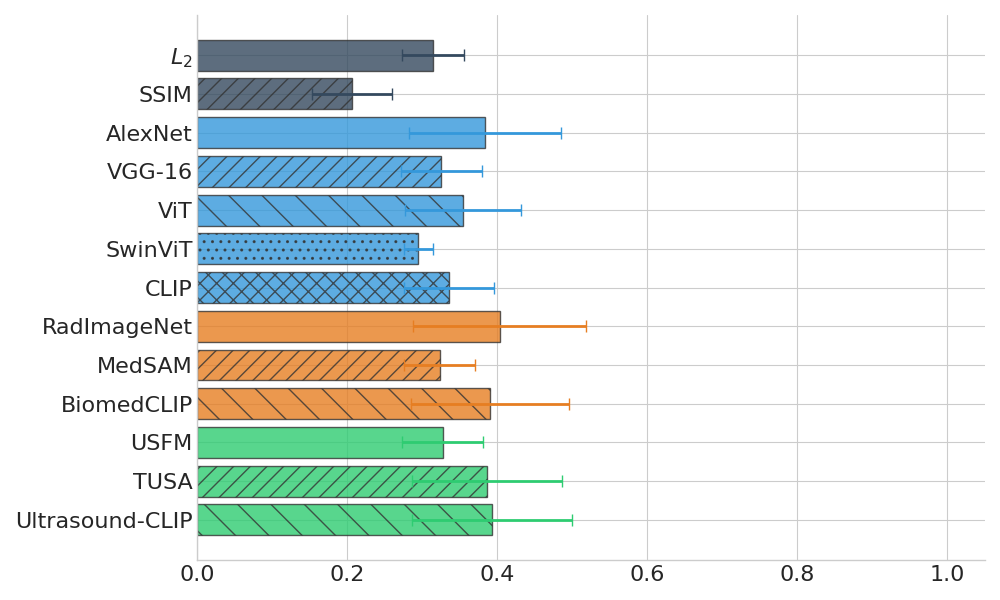} &
        \includegraphics[width=0.5\textwidth, valign=m]{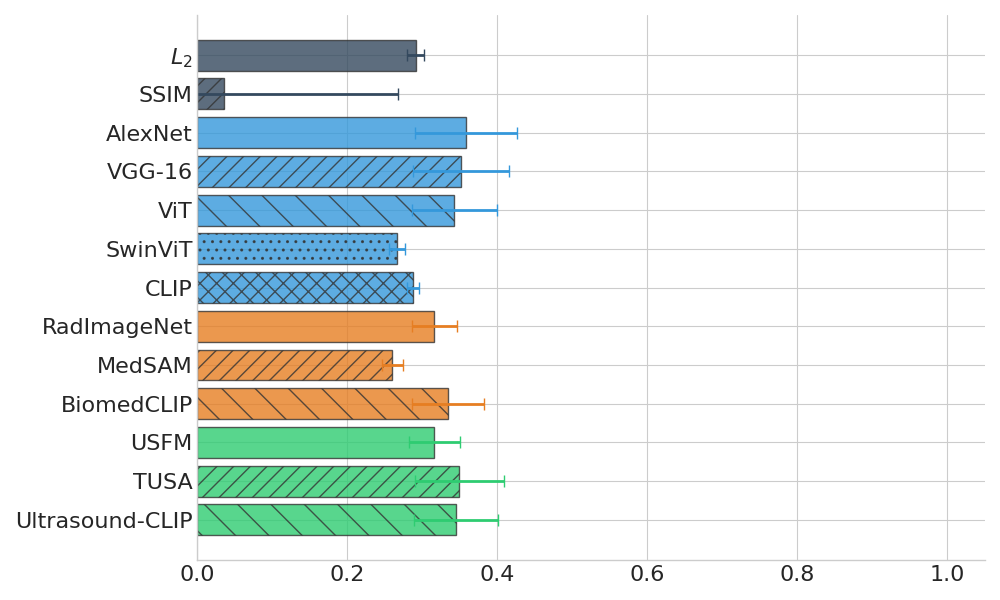} \\
    \end{tabular}
    \includegraphics[width=0.8\textwidth]{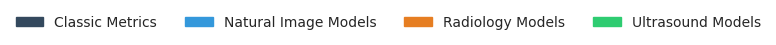}
    \caption{
        Spearman correlations of supervised models on CAMUS and EchoNet datasets under ultrasound augmentation.
        In the view classification task (a) correlation is calculated between the log-softmax confidence score of EchoPrime's view classifier and each distance metric, averaged across augmentations and frames.
        For segmentation (b), the correlation is calculated between UltraSam's dice score and and each of the distance metrics, averaged across augmentations and frames
    }
    \label{fig:figure3}
\end{figure*}
}

% --- Figure: INR Visual Comparison (Fig 2) ---
\newcommand{\figInrFollicles}{
\begin{figure*}[tb]
    \centering
    \setlength{\fboxrule}{3pt}
    \setlength{\fboxsep}{0pt}
    \setlength{\tabcolsep}{4pt}
    \renewcommand{\arraystretch}{0.1}
    \begin{minipage}{\textwidth}
        \centering
        \includegraphics[width=0.2\textwidth]{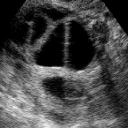} \\
        Ground Truth \\[0.5em]
        
        \resizebox{0.8\textwidth}{!}{
            \begin{tabular}{@{}cccc@{}}
                \fcolorbox{ClassicalCol}{white}{\includegraphics[width=0.21\textwidth]{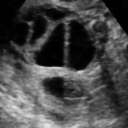}} &
                \fcolorbox{ClassicalCol}{white}{\includegraphics[width=0.21\textwidth]{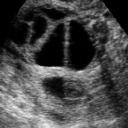}} &
                \fcolorbox{ImageNetCol}{white}{\includegraphics[width=0.21\textwidth]{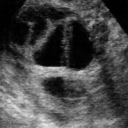}} &
                \fcolorbox{ImageNetCol}{white}{\includegraphics[width=0.21\textwidth]{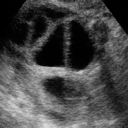}} \\[0.4em]
                $L_2$ & SSIM & AlexNet & VGG-16 \\[0.5em]
        
                \fcolorbox{ImageNetCol}{white}{\includegraphics[width=0.21\textwidth]{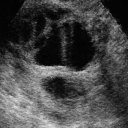}} &
                \fcolorbox{ImageNetCol}{white}{\includegraphics[width=0.21\textwidth]{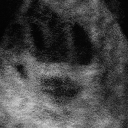}} &
                \fcolorbox{ImageNetCol}{white}{\includegraphics[width=0.21\textwidth]{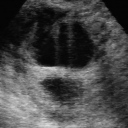}} &
                \fcolorbox{RadiologyCol}{white}{\includegraphics[width=0.21\textwidth]{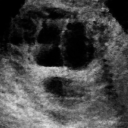}} \\[0.4em]
                ViT & SwinViT & CLIP & RadImageNet \\[0.5em]
        
                \fcolorbox{RadiologyCol}{white}{\includegraphics[width=0.21\textwidth]{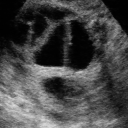}} &
                \fcolorbox{UltrasoundCol}{white}{\includegraphics[width=0.21\textwidth]{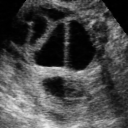}} &
                \fcolorbox{UltrasoundCol}{white}{\includegraphics[width=0.21\textwidth]{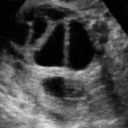}} &
                \fcolorbox{UltrasoundCol}{white}{\includegraphics[width=0.21\textwidth]{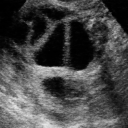}} \\[0.4em]
                BiomedCLIP & USFM & TUSA & Ultrasound-CLIP \\
            \end{tabular}
            }
            \includegraphics[width=0.8\textwidth]{legend.png}
        \end{minipage}
    \caption{
        Implicit Neural Representation (INR) visual comparison on the follicles dataset. Models are grouped and color-coded by their pre-training domain: Classical (dark blue), Natural Images (light blue), Radiology (orange), and Ultrasound (green).
    }
    \label{fig:inr_follicles}
\end{figure*}
}

% --- Figure: EchoGains Augmentations (Supplementary Fig S1) ---
\makeatletter
\newcommand{\figEchoGains}{
\begin{center}
    \vspace{-0.5em}
    \setlength{\tabcolsep}{0pt}
    \renewcommand{\arraystretch}{0.1}
    \begin{tabular}{@{}c@{}cccc@{}}

        % ---------------- Row 0 : Depth ----------------
        \adjustbox{valign=m}{\rotatebox{90}{\small Depth}} &
        \includegraphics[width=0.19\textwidth, valign=m]{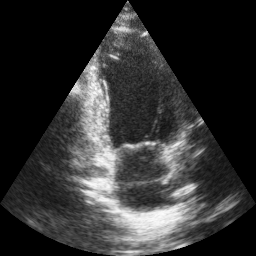} &
        \includegraphics[width=0.19\textwidth, valign=m]{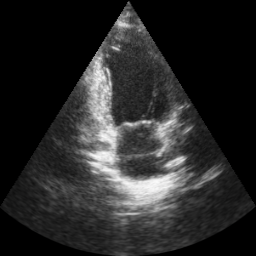} &
        \includegraphics[width=0.19\textwidth, valign=m]{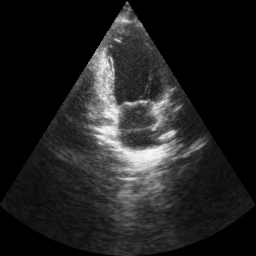} &
        \includegraphics[width=0.19\textwidth, valign=m]{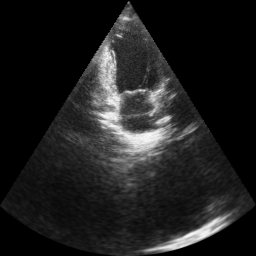} \\

        % ---------------- Row 1 : Rotation ----------------
        \adjustbox{valign=m}{\rotatebox{90}{\small Rotation}} &
        \includegraphics[width=0.19\textwidth, valign=m]{00/original.png} &
        \includegraphics[width=0.19\textwidth, valign=m]{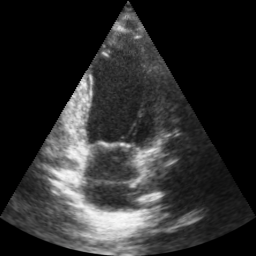} &
        \includegraphics[width=0.19\textwidth, valign=m]{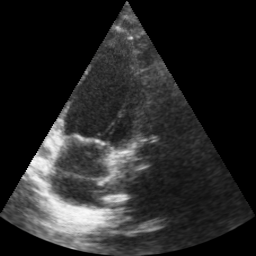} &
        \includegraphics[width=0.19\textwidth, valign=m]{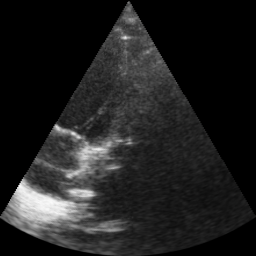} \\

        % ---------------- Row 2 : Sector Width ----------------
        \adjustbox{valign=m}{\rotatebox{90}{\small \begin{tabular}{@{}c@{}}Sector\\Width\end{tabular}}} &
        \includegraphics[width=0.19\textwidth, valign=m]{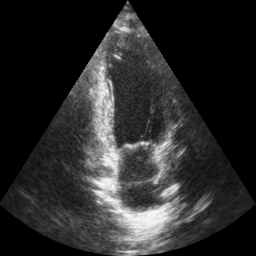} &
        \includegraphics[width=0.19\textwidth, valign=m]{00/original.png} &
        \includegraphics[width=0.19\textwidth, valign=m]{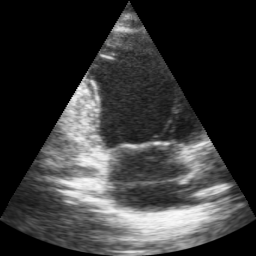} &
        \includegraphics[width=0.19\textwidth, valign=m]{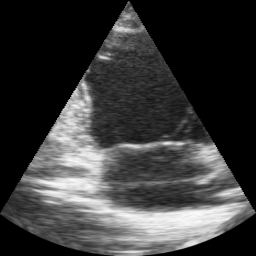} \\

        % ---------------- Row 3 : Translation ----------------
        \adjustbox{valign=m}{\rotatebox{90}{\small Translation}} &
        \includegraphics[width=0.19\textwidth, valign=m]{00/original.png} &
        \includegraphics[width=0.19\textwidth, valign=m]{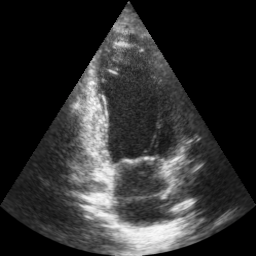} &
        \includegraphics[width=0.19\textwidth, valign=m]{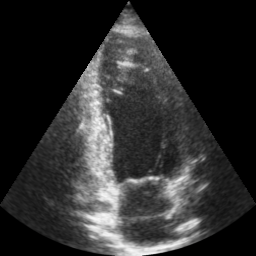} &
        \includegraphics[width=0.19\textwidth, valign=m]{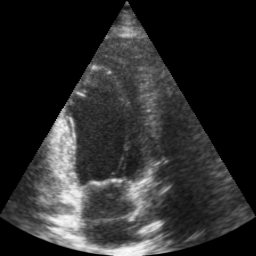} \\

    \end{tabular}

    \vspace{0.3em}

    % ---------------- Bottom Arrow ----------------
    \begin{tikzpicture}
        \draw[->, line width=1.5pt]
            (0,0) -- (8,0)
            node[midway, below=0.15cm] {\small Increased augmentation strength};
    \end{tikzpicture}

    \vspace{0.5em}
    \def\@captype{figure}
    \caption{
        Ultrasound augmentations provided by EchoGains.
        The pre-trained EchoGains diffusion model is used to induce realistic controlled perturbation in B-mode images.
    }
    \label{fig:echogain}
\end{center}
}

% --- Figure: UltraSam Predictions (Supplementary Fig S2) ---
\newcommand{\figUltrasamPredictions}{
\begin{center}
    \setlength{\tabcolsep}{0pt}
    \renewcommand{\arraystretch}{0.1}
    \begin{tabular}{@{}l@{\hspace{4pt}}ccccc@{}}
        \adjustbox{valign=m}{\textbf{(a)}} &
        \includegraphics[width=0.19\textwidth, valign=m]{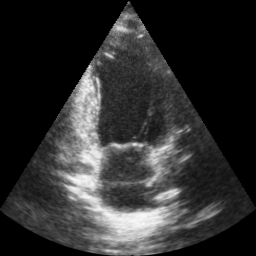} &
        \includegraphics[width=0.19\textwidth, valign=m]{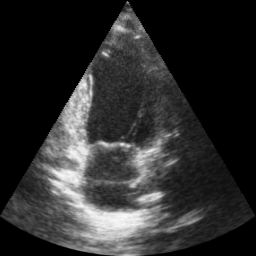} &
        \includegraphics[width=0.19\textwidth, valign=m]{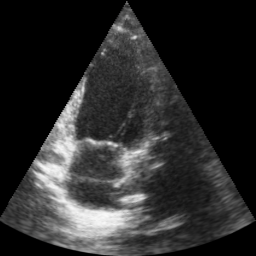} &
        \includegraphics[width=0.19\textwidth, valign=m]{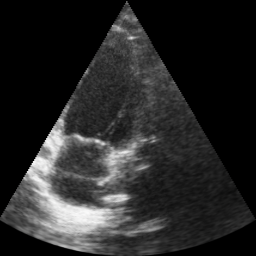} &
        \includegraphics[width=0.19\textwidth, valign=m]{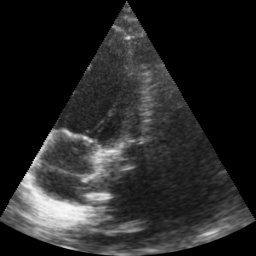} \\
        
        \adjustbox{valign=m}{\textbf{(b)}} &
        \includegraphics[width=0.19\textwidth, valign=m]{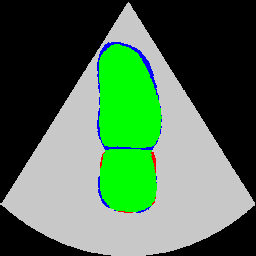} &
        \includegraphics[width=0.19\textwidth, valign=m]{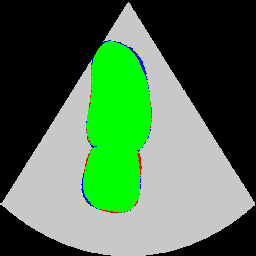} &
        \includegraphics[width=0.19\textwidth, valign=m]{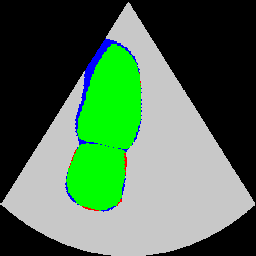} &
        \includegraphics[width=0.19\textwidth, valign=m]{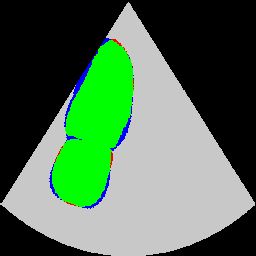} &
        \includegraphics[width=0.19\textwidth, valign=m]{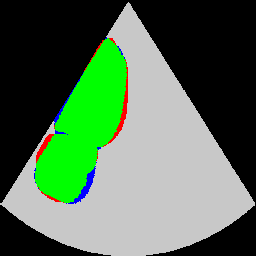} \\

        & \multicolumn{5}{@{}c@{}}{\includegraphics[width=0.8\textwidth]{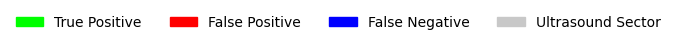}} \\

        \adjustbox{valign=m}{\textbf{(c)}} &
        \includegraphics[width=0.19\textwidth, valign=m]{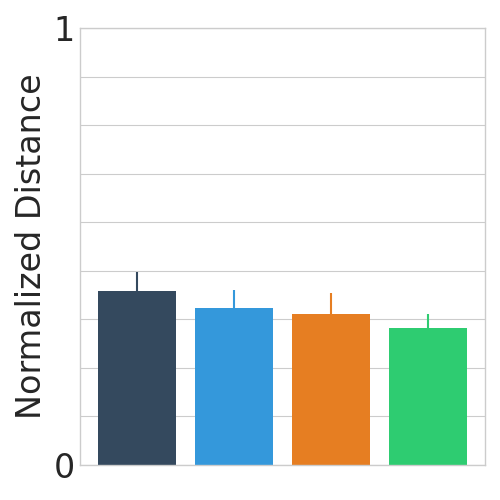} &
        \includegraphics[width=0.19\textwidth, valign=m]{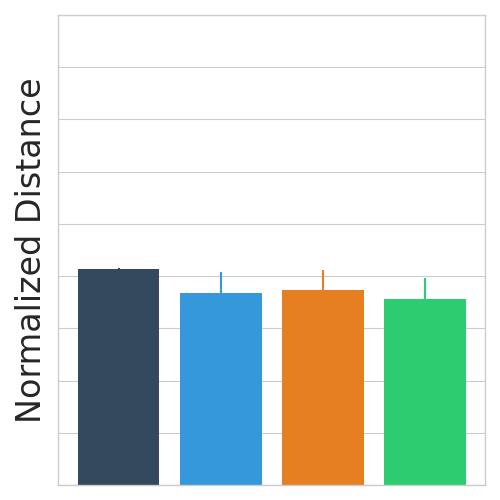} &
        \includegraphics[width=0.19\textwidth, valign=m]{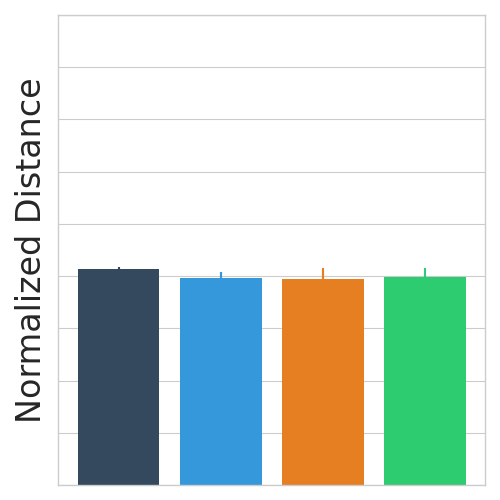} &
        \includegraphics[width=0.19\textwidth, valign=m]{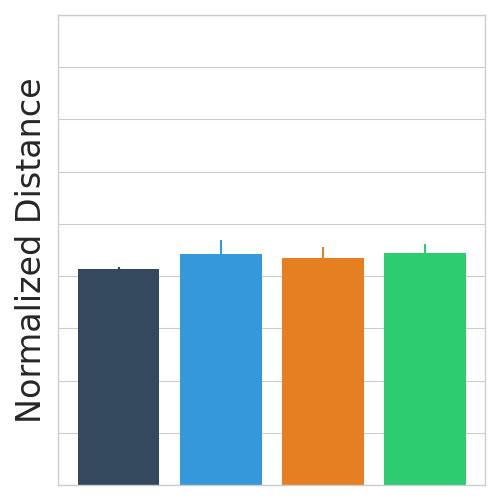} &
        \includegraphics[width=0.19\textwidth, valign=m]{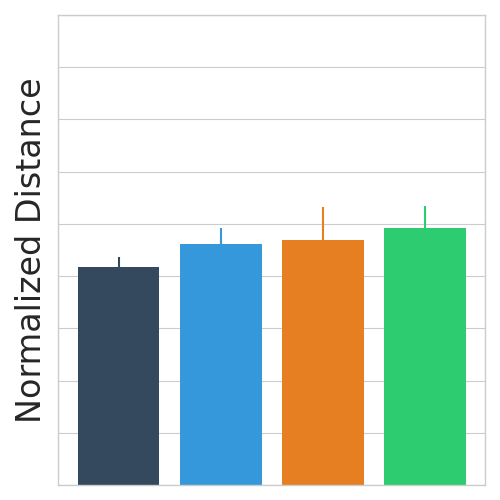} \\
    \end{tabular}
    \vspace{0.3em}
    \includegraphics[width=0.8\textwidth]{legend.png}
    
    \begin{tikzpicture}
        \draw[->, line width=1.5pt]
        (0,0) -- (8,0)
        node[midway, below=0.15cm] {\small Increased augmentation strength};
    \end{tikzpicture}
    
    \vspace{0.5em}
    \def\@captype{figure}
    \caption{
        UltraSam segmentation of left ventricle and atrium.
        As increasingly stronger rotation is applied to the B-mode image (a), the model still recognizes the heart chambers but becomes more error prone (b).
        This corresponds to increased perturbation captured by distance metrics (c).
        }
        \label{fig:ultrasam_predictions}
\end{center}
}
\makeatother

% Tables for UltraPIPS paper

\newcommand{\tabStats}{
\begin{table}[t]
    \centering
    \caption{Wilcoxon test of LPIPS computed with ultrasound models vs. comparison groups.}
    \label{tab:wilcoxon}
    \begin{tabular}{lcc}
        \toprule
        \textbf{Null Hypothesis} & \textbf{Classification} & \textbf{Segmentation} \\
        \midrule
        $\rho_{\text{US}} \leq \rho_{\text{Classical}}$  & \hspace{1cm} $7.8 \times 10^{-3}$ \hspace{1cm} & $7.8 \times 10^{-3}$ \\
        $\rho_{\text{US}} \leq \rho_{\text{Natural Images}}$   & \hspace{1cm} 0.016 \hspace{1cm} & 0.039 \\
        $\rho_{\text{US}} \leq \rho_{\text{Radiology}}$  & \hspace{1cm} 0.078 \hspace{1cm} & 0.078 \\
        \bottomrule
    \end{tabular}
\end{table}
}

\newcommand{\tabINRGrouped}{
\begin{table}[t]
\centering
\caption{Image reconstruction metrics of INRs trained with each group of distance metrics. For each reconstruction metric, the best score for each dataset among the groups of distance metrics is displayed in \textbf{bold}.}
\label{tab:inr_grouped}
\resizebox{\textwidth}{!}{\begin{tabular}{l|cccc|cccc}
\toprule
 & \multicolumn{4}{c|}{USOVA} & \multicolumn{4}{c}{MicroSegNet} \\
 & $L_2$ & Natural Images & Radiology & Ultrasound & $L_2$ & Natural Images & Radiology & Ultrasound \\
\midrule
$\uparrow$ SSIM & \textbf{0.76 $\pm$ 0.09} & 0.45 $\pm$ 0.17 & 0.47 $\pm$ 0.13 & 0.70 $\pm$ 0.13 & \textbf{0.65 $\pm$ 0.04} & 0.36 $\pm$ 0.13 & 0.39 $\pm$ 0.09 & 0.55 $\pm$ 0.10 \\
$\downarrow$ HFEN & \textbf{0.34 $\pm$ 0.04} & 1.12 $\pm$ 0.45 & 1.06 $\pm$ 0.27 & 0.46 $\pm$ 0.15 & \textbf{0.37 $\pm$ 0.04} & 1.41 $\pm$ 0.42 & 1.36 $\pm$ 0.25 & 0.71 $\pm$ 0.29 \\
$\downarrow$ GLCM Contrast & 2.84 $\pm$ 1.92 & 1.81 $\pm$ 1.43 & \textbf{1.79 $\pm$ 1.37} & 1.96 $\pm$ 1.89 & 5.42 $\pm$ 1.84 & 3.61 $\pm$ 2.07 & 3.87 $\pm$ 1.83 & \textbf{3.24 $\pm$ 2.86} \\
$\downarrow$ GLCM Dissimilarity & 0.45 $\pm$ 0.26 & 0.24 $\pm$ 0.18 & \textbf{0.23 $\pm$ 0.16} & 0.29 $\pm$ 0.26 & 0.62 $\pm$ 0.21 & \textbf{0.28 $\pm$ 0.18} & 0.36 $\pm$ 0.19 & 0.33 $\pm$ 0.32 \\
$\downarrow$ GLCM Homogeneity & 0.07 $\pm$ 0.04 & 0.04 $\pm$ 0.04 & \textbf{0.03 $\pm$ 0.02} & 0.04 $\pm$ 0.04 & 0.06 $\pm$ 0.03 & 0.04 $\pm$ 0.03 & \textbf{0.03 $\pm$ 0.02} & \textbf{0.03 $\pm$ 0.03} \\
$\downarrow$ GLCM Energy & 0.02 $\pm$ 0.02 & 0.04 $\pm$ 0.05 & 0.02 $\pm$ 0.02 & \textbf{0.01 $\pm$ 0.01} & 0.02 $\pm$ 0.01 & 0.04 $\pm$ 0.03 & \textbf{0.01 $\pm$ 0.01} & \textbf{0.01 $\pm$ 0.01} \\
$\downarrow$ GLCM Correlation & 0.05 $\pm$ 0.04 & \textbf{0.03 $\pm$ 0.02} & \textbf{0.03 $\pm$ 0.02} & \textbf{0.03 $\pm$ 0.03} & 0.09 $\pm$ 0.02 & \textbf{0.05 $\pm$ 0.02} & 0.06 $\pm$ 0.02 & 0.06 $\pm$ 0.04 \\
\bottomrule
\end{tabular}}
\end{table}
}

% --- Supplementary Tables ---

\newcommand{\tabClassificationDetailed}{
\begin{table}[!htbp]
    \caption{
        Spearman correlation of EchoPrime view classifier confidence with distance metrics on augmented data.
        For each augmentation type, the augmentation is applied with increased strength.
        Correlations are measured between view classifier log-softmax output at the correct view class and each of the candidate distance metrics, averaged across frames.
        In each column, the highest correlation is displayed in \textbf{bold}.
        }
        \label{tab:S1}
        \centering\resizebox{\textwidth}{!}{\begin{tabular}{c|cccc|cccc|c}
            \toprule
            & \multicolumn{4}{c|}{CAMUS} & \multicolumn{4}{c|}{EchoNet} & Overall \\
            Model & Depth & Rotation & Sector Width & Translation & Depth & Rotation & Sector Width & Translation &  \\
    \midrule
    $L_2$ & 0.57 $\pm$ 0.25 & 0.36 $\pm$ 0.29 & 0.58 $\pm$ 0.27 & 0.26 $\pm$ 0.28 & 0.50 $\pm$ 0.27 & 0.40 $\pm$ 0.29 & 0.53 $\pm$ 0.27 & 0.37 $\pm$ 0.25 & 0.45 $\pm$ 0.28 \\
    SSIM & 0.42 $\pm$ 0.26 & 0.20 $\pm$ 0.28 & 0.51 $\pm$ 0.28 & 0.18 $\pm$ 0.25 & 0.24 $\pm$ 0.27 & 0.09 $\pm$ 0.28 & 0.23 $\pm$ 0.27 & 0.02 $\pm$ 0.22 & 0.20 $\pm$ 0.27 \\
    \midrule
    AlexNet & 0.58 $\pm$ 0.25 & 0.45 $\pm$ 0.26 & $\mathbf{0.62 \pm 0.25}$ & 0.27 $\pm$ 0.27 & 0.54 $\pm$ 0.28 & 0.51 $\pm$ 0.29 & 0.56 $\pm$ 0.27 & 0.39 $\pm$ 0.25 & 0.49 $\pm$ 0.28 \\
    VGG-16 & 0.54 $\pm$ 0.25 & 0.37 $\pm$ 0.27 & 0.61 $\pm$ 0.26 & 0.25 $\pm$ 0.27 & $\mathbf{0.55 \pm 0.27}$ & 0.52 $\pm$ 0.29 & 0.57 $\pm$ 0.27 & 0.39 $\pm$ 0.25 & 0.48 $\pm$ 0.28 \\
    ViT & 0.52 $\pm$ 0.26 & 0.38 $\pm$ 0.26 & 0.62 $\pm$ 0.26 & 0.24 $\pm$ 0.27 & 0.53 $\pm$ 0.27 & 0.50 $\pm$ 0.28 & 0.55 $\pm$ 0.27 & 0.38 $\pm$ 0.25 & 0.47 $\pm$ 0.28 \\
    SwinViT & 0.44 $\pm$ 0.26 & 0.34 $\pm$ 0.29 & 0.57 $\pm$ 0.27 & 0.13 $\pm$ 0.24 & 0.35 $\pm$ 0.25 & 0.39 $\pm$ 0.29 & 0.49 $\pm$ 0.27 & 0.27 $\pm$ 0.23 & 0.37 $\pm$ 0.28 \\
    CLIP & 0.51 $\pm$ 0.26 & 0.39 $\pm$ 0.27 & 0.57 $\pm$ 0.27 & 0.23 $\pm$ 0.26 & 0.48 $\pm$ 0.27 & 0.41 $\pm$ 0.30 & 0.48 $\pm$ 0.26 & 0.35 $\pm$ 0.24 & 0.43 $\pm$ 0.28 \\
    \midrule
    RadImageNet & $\mathbf{0.61 \pm 0.26}$ & $\mathbf{0.49 \pm 0.26}$ & 0.62 $\pm$ 0.26 & 0.28 $\pm$ 0.28 & 0.50 $\pm$ 0.26 & 0.48 $\pm$ 0.29 & 0.48 $\pm$ 0.27 & 0.35 $\pm$ 0.25 & 0.47 $\pm$ 0.29 \\
    MedSAM & 0.58 $\pm$ 0.26 & 0.36 $\pm$ 0.30 & 0.58 $\pm$ 0.27 & 0.24 $\pm$ 0.28 & 0.46 $\pm$ 0.27 & 0.31 $\pm$ 0.29 & 0.49 $\pm$ 0.27 & 0.34 $\pm$ 0.25 & 0.42 $\pm$ 0.29 \\
    BiomedCLIP & 0.59 $\pm$ 0.26 & 0.48 $\pm$ 0.26 & 0.61 $\pm$ 0.27 & 0.26 $\pm$ 0.28 & 0.51 $\pm$ 0.27 & 0.50 $\pm$ 0.29 & 0.55 $\pm$ 0.27 & 0.38 $\pm$ 0.25 & 0.48 $\pm$ 0.29 \\
    \midrule
    USFM & 0.56 $\pm$ 0.25 & 0.37 $\pm$ 0.28 & 0.56 $\pm$ 0.26 & 0.27 $\pm$ 0.28 & 0.54 $\pm$ 0.27 & 0.42 $\pm$ 0.29 & 0.53 $\pm$ 0.27 & 0.37 $\pm$ 0.25 & 0.45 $\pm$ 0.28 \\
    TUSA & 0.59 $\pm$ 0.26 & 0.48 $\pm$ 0.25 & 0.60 $\pm$ 0.27 & 0.27 $\pm$ 0.28 & 0.53 $\pm$ 0.27 & $\mathbf{0.53 \pm 0.29}$ & $\mathbf{0.57 \pm 0.26}$ & $\mathbf{0.40 \pm 0.25}$ & $\mathbf{0.50 \pm 0.28}$ \\
    Ultrasound-CLIP & 0.58 $\pm$ 0.25 & 0.48 $\pm$ 0.25 & 0.62 $\pm$ 0.26 & $\mathbf{0.28 \pm 0.28}$ & 0.50 $\pm$ 0.27 & 0.51 $\pm$ 0.29 & 0.55 $\pm$ 0.26 & 0.39 $\pm$ 0.25 & 0.49 $\pm$ 0.28 \\
    \bottomrule
\end{tabular}}
\end{table}
}

\newcommand{\tabSegmentationDetailed}{
\begin{table}[!htbp]
\caption{
        Spearman correlation of UltraSam dice scores with distance metrics on augmented data.
        For each augmentation type, the augmentation is applied with increased strength.
        Correlations are measured between dice score and each of the candidate distance metrics, averaged across frames.
        In each column, the highest correlation is displayed in \textbf{bold}.
        }
\label{tab:S2}
\centering\resizebox{\textwidth}{!}{\begin{tabular}{c|cccc|cccc|c}
\toprule
 & \multicolumn{4}{c|}{CAMUS} & \multicolumn{4}{c|}{EchoNet} & Overall \\
Model & Depth & Rotation & Sector Width & Translation & Depth & Rotation & Sector Width & Translation &  \\
\midrule
$L_2$ & 0.04 $\pm$ 0.25 & 0.46 $\pm$ 0.29 & 0.45 $\pm$ 0.28 & 0.38 $\pm$ 0.25 & 0.10 $\pm$ 0.25 & 0.39 $\pm$ 0.30 & 0.31 $\pm$ 0.28 & 0.36 $\pm$ 0.25 & 0.30 $\pm$ 0.28 \\
SSIM & 0.01 $\pm$ 0.22 & 0.20 $\pm$ 0.27 & 0.39 $\pm$ 0.28 & 0.25 $\pm$ 0.24 & 0.06 $\pm$ 0.23 & 0.04 $\pm$ 0.30 & 0.14 $\pm$ 0.26 & 0.01 $\pm$ 0.22 & 0.12 $\pm$ 0.26 \\
\midrule
AlexNet & 0.04 $\pm$ 0.25 & 0.67 $\pm$ 0.26 & $\mathbf{0.51 \pm 0.28}$ & $\mathbf{0.40 \pm 0.25}$ & $\mathbf{0.13 \pm 0.27}$ & $\mathbf{0.57 \pm 0.29}$ & $\mathbf{0.35 \pm 0.29}$ & $\mathbf{0.39 \pm 0.25}$ & $\mathbf{0.37 \pm 0.29}$ \\
VGG-16 & 0.02 $\pm$ 0.23 & 0.50 $\pm$ 0.28 & 0.48 $\pm$ 0.28 & 0.35 $\pm$ 0.24 & $\mathbf{0.13 \pm 0.27}$ & 0.55 $\pm$ 0.29 & $\mathbf{0.35 \pm 0.28}$ & 0.38 $\pm$ 0.25 & 0.34 $\pm$ 0.28 \\
ViT & 0.02 $\pm$ 0.24 & 0.59 $\pm$ 0.27 & 0.48 $\pm$ 0.28 & 0.37 $\pm$ 0.25 & 0.12 $\pm$ 0.26 & 0.54 $\pm$ 0.30 & 0.34 $\pm$ 0.28 & 0.37 $\pm$ 0.25 & 0.35 $\pm$ 0.28 \\
SwinViT & 0.04 $\pm$ 0.24 & 0.52 $\pm$ 0.29 & 0.43 $\pm$ 0.27 & 0.27 $\pm$ 0.24 & 0.10 $\pm$ 0.24 & 0.42 $\pm$ 0.30 & 0.30 $\pm$ 0.27 & 0.25 $\pm$ 0.23 & 0.28 $\pm$ 0.28 \\
CLIP & 0.03 $\pm$ 0.24 & 0.58 $\pm$ 0.27 & 0.45 $\pm$ 0.29 & 0.33 $\pm$ 0.24 & 0.10 $\pm$ 0.26 & 0.45 $\pm$ 0.30 & 0.29 $\pm$ 0.27 & 0.31 $\pm$ 0.23 & 0.31 $\pm$ 0.28 \\
\midrule
RadImageNet & 0.03 $\pm$ 0.26 & $\mathbf{0.75 \pm 0.24}$ & 0.50 $\pm$ 0.29 & $\mathbf{0.40 \pm 0.25}$ & 0.11 $\pm$ 0.25 & 0.53 $\pm$ 0.29 & 0.29 $\pm$ 0.28 & 0.34 $\pm$ 0.25 & 0.36 $\pm$ 0.29 \\
MedSAM & 0.03 $\pm$ 0.25 & 0.48 $\pm$ 0.29 & 0.46 $\pm$ 0.29 & 0.38 $\pm$ 0.25 & 0.09 $\pm$ 0.25 & 0.34 $\pm$ 0.29 & 0.28 $\pm$ 0.28 & 0.33 $\pm$ 0.25 & 0.29 $\pm$ 0.28 \\
BiomedCLIP & 0.04 $\pm$ 0.26 & 0.73 $\pm$ 0.24 & 0.47 $\pm$ 0.28 & 0.39 $\pm$ 0.25 & 0.10 $\pm$ 0.26 & 0.55 $\pm$ 0.30 & 0.33 $\pm$ 0.28 & 0.36 $\pm$ 0.24 & 0.36 $\pm$ 0.29 \\
\midrule
USFM & 0.04 $\pm$ 0.25 & 0.53 $\pm$ 0.28 & 0.45 $\pm$ 0.28 & 0.37 $\pm$ 0.25 & 0.12 $\pm$ 0.26 & 0.46 $\pm$ 0.30 & 0.32 $\pm$ 0.28 & 0.37 $\pm$ 0.25 & 0.32 $\pm$ 0.28 \\
TUSA & $\mathbf{0.05 \pm 0.26}$ & 0.73 $\pm$ 0.24 & 0.47 $\pm$ 0.28 & $\mathbf{0.40 \pm 0.26}$ & 0.11 $\pm$ 0.26 & 0.56 $\pm$ 0.30 & 0.34 $\pm$ 0.28 & $\mathbf{0.39 \pm 0.25}$ & $\mathbf{0.37 \pm 0.29}$ \\
Ultrasound-CLIP & 0.04 $\pm$ 0.25 & 0.74 $\pm$ 0.23 & 0.48 $\pm$ 0.28 & 0.39 $\pm$ 0.25 & 0.10 $\pm$ 0.26 & 0.56 $\pm$ 0.30 & 0.33 $\pm$ 0.28 & 0.38 $\pm$ 0.24 & $\mathbf{0.37 \pm 0.29}$ \\
\bottomrule
\end{tabular}}
\end{table}
}

\title{UltraPIPS: Improving model perception in B-mode ultrasound with foundation models} 
\titlerunning{UltraPIPS}

\author{Tal Grutman\orcidID{0009-0000-8192-4958} \and
Tali Ilovitsh\orcidID{0000-0001-6215-0299}}
\authorrunning{T. Grutman, T. Ilovitsh}
\institute{School of Biomedical Engineering, Tel Aviv University, Tel Aviv, Israel\\
\email{ilovitsh@tauex.tau.ac.il}}

\maketitle

\begin{tikzpicture}[remember picture,overlay]
  \node[anchor=north east, text=black, font=\footnotesize] at ([xshift=-2.2cm, yshift=-1.3cm]current page.north east) {Accepted to MICCAI ASMUS 2026};
\end{tikzpicture}

\begin{abstract}
In medical imaging, it is common to use learned perceptual image patch similarity (LPIPS) to compare images semantically in feature space. Although backbones pretrained on natural images are widely used for LPIPS computation, B-mode ultrasound images possess distinct speckle patterns and acoustic-specific image statistics that are fundamentally different from natural images and even from other images in radiology. Consequently, we propose that domain-specific models are needed to measure perceptual similarity in ultrasound data, a finding which is not necessarily the case for other imaging modalities. We compare LPIPS metrics across downstream tasks like classification, segmentation and reconstruction using natural image, medical generalist and ultrasound backbone models and show that selection of LPIPS backbone is a non-trivial design choice. In particular, the ultrasound backbone models were more correlated with downstream performance of supervised models than classical and natural image models, and optimization of the LPIPS loss with an ultrasound backbone achieved a strong balance between reconstruction quality and realism. Our code is available at \url{https://github.com/talg2324/UltraPIPS} and introduces the UltraPIPS library, a set of LPIPS metrics based on the open-source foundation models analyzed in this paper.
\keywords{Ultrasound \and Foundation models \and Perceptual similarity \and Computer vision}
\end{abstract}

\section{Introduction}
\label{sec:intro}

Learned perceptual image patch similarity (LPIPS) \cite{lpips} is an important perceptual metric that uses a pretrained image encoder to measure distance between image patches in the encoder's feature space. The advantage of this approach is that the resulting metric is grounded by the downstream task on which the encoder was trained, with the deep features extracted by the metric incorporating rich context with which to perceive the difference between patches. LPIPS is highly correlated to downstream classification and detection tasks, suggesting that it is a soft middle-ground between human perception and performance of deep models. One reason for this is that the LPIPS backbone models were pretrained in self-supervised or supervised learning tasks that require global context to be embedded in the feature space. This provides a natural improvement from metrics like standard $L_2$ or structural similarity index measure (SSIM) \cite{ssim}. Since it was first introduced, loss functions and evaluation criteria based on LPIPS have become well-established in image synthesis \cite{ldm}, implicit neural representations \cite{grutman2025}, and image reconstruction \cite{li2025,stevens2025}. LPIPS is widely used in computer vision, and has been extended into medical imaging \cite{grutman2026,cahan2025} often using the RadImageNet \cite{radimagenet} backbone.

There is no consensus on the importance of the backbone model to the effectiveness of the LPIPS metric, and the choice of backbone is not well studied in medical imaging. Backbone choice is conceptually related to transfer learning, where several studies found that in-domain training or pretraining added marginal performance or even performed worse than models pretrained on ImageNet \cite{yuan2025,zu2025}. More specifically in the context of LPIPS, it has been shown that in-domain LPIPS backbones did not improve MRI reconstruction \cite{adamson2025} compared to ImageNet ones. These findings challenge the necessity of in-domain pretraining, but other notable works have found that in-domain medical image models learn vastly different features with much higher relevance in comparison to ImageNet-based models \cite{radimagenet,transfusion}.

Some of the discrepancies in results can be attributed to the various domains within medical imaging, each possessing highly unique spatial resolutions, dynamic range, and signal-to-noise ratio. Ultrasound is well-known to be an especially difficult imaging domain, and foundation models trained exclusively on ultrasound images have been shown to outperform general-purpose radiology models \cite{grutman2026,ultrasam} on a variety of tasks. LPIPS is particularly important in ultrasound models because deep backbones can accommodate the unique speckle characteristics common to B-mode ultrasound, which are quickly lost with pixel-wise losses like MSE \cite{li2025} despite carrying important diagnostic meaning \cite{wagner1983}. Thus, LPIPS is widely used for training and evaluation of reconstruction-oriented ultrasound models \cite{grutman2025,li2025,stevens2025}. We hypothesize that ultrasound reconstruction tasks like these can be improved by foundation models built for ultrasound images to power their LPIPS metrics, as they are better accustomed to the unique properties of ultrasound.

In this work, we investigate the importance of ultrasound foundation models in perceptual similarity of B-mode ultrasound images compared to models trained on natural or general medical images. We utilize open-source, pretrained models and datasets freely available on the web to challenge the usage of non-ultrasound backbones in LPIPS metrics by studying the downstream relationship of such LPIPS distances with classification, segmentation, and image reconstruction. We test feasible candidates for LPIPS metrics in ultrasound and open-source the UltraPIPS library which contains multiple backbones and tests for perceptual ultrasound metrics.

\section{Experiments}

Several cross-domain encoder candidates exist for an ultrasound perceptual metric. Models like RadImageNet \cite{radimagenet} and MedSAM \cite{medsam} which were trained on large-scale medical image datasets (including but not exclusive to ultrasound) could potentially provide useful features based on their deep understanding of anatomy. This is especially true for MedSAM, which was trained specifically to segment diagnostically important information.

Alternatively, ultrasound foundation models can provide an ultrasound-first metric that can accommodate for the discrepancies between ultrasound and other modalities. To produce a useful model for perceptual metrics, such foundation models must be trained on a wide variety of organs with exposure to various imaging acquisition parameters. Several open-source candidates exist. Ultrasound Foundation Model (USFM) \cite{usfm} is a ViT-b encoder that was trained in a masked auto-encoding framework. Texture ultrasound semantic analysis (TUSA) \cite{grutman2026} used a learned dictionary of cross-anatomy texture kernels to reconstruct images, producing a SwinViT \cite{swinvit} backbone grounded in B-mode image texture characteristics. Ultrasound-CLIP \cite{ultrasound-clip} is an ultrasound foundation model trained to extract global semantic features from ultrasound images based on corresponding text reports. These models are attractive candidates for an LPIPS backbone, and we hypothesize that with increased training grounded in the properties of ultrasound, they will outperform natural image or radiology models as perceptual metrics. We are not aware of any modern CNN-based ultrasound foundation model, limiting our analysis to ViT and SwinViT variants. We analyzed ImageNet variants of ViT and SwinViT to control for the impact of training data on these architectures. Similarly, we added the original CLIP \cite{clip} and the medical generalist BiomedCLIP \cite{biomedCLIP} models to our analysis to isolate the effect of the training data compared to Ultrasound-CLIP and training strategy compared to self-supervised and classification models. $L_2$ and SSIM serve as classical metric baselines; candidate LPIPS backbones are grouped by pre-training domain in table~1.

\begin{table}[h]
\centering
\caption{Candidate LPIPS backbone models grouped by pre-training domain.}
\label{tab:models}
\begin{tabular}{ccc}
\toprule
\textbf{Natural Image} & \textbf{Radiology} & \textbf{Ultrasound} \\
\midrule
AlexNet    & RadImageNet & USFM            \\
VGG-16     & MedSAM      & TUSA            \\
ViT        & BiomedCLIP  & Ultrasound-CLIP \\
SwinViT    &             &                 \\
CLIP       &             &                 \\
\bottomrule
\end{tabular}
\end{table}

The MONAI library was used to calculate the RadImageNet LPIPS, and the original LPIPS \cite{lpips} implementation was used for AlexNet and VGG-16. For ViT/SwinViT variants, features were extracted at the final layer to produce a feature vector, in line with the LPIPS and MONAI implementations. The LPIPS loss is the $L_2$ loss between channel-normalized feature vectors of two images.

Experiments were run on a Kubernetes cluster with the Run:ai (Tel-Aviv, Israel) resource manager on an NVIDIA RTX A5000 GPU (NVIDIA Corporation, Santa Clara, CA, USA) using PyTorch.

\subsection{Supervised downstream tasks}
We used the EchoGains \cite{echogains} library to induce ultrasound-based augmentations on echocardiogram B-mode images. This unique and recent augmentation strategy presents a realistic scenario for image degradation in clinical context. EchoGains achieves this by performing geometric warping of ultrasound images, masking out pixels outside the original transducer field-of-view, and using a diffusion model to inpaint the missing pixels in the ultrasound sector (see Fig. S1 in supplementary). It offers augmentations of the imaging depth, sector width, probe translation and rotation.

\paragraph{Cardiac View Classification} The coarse view classifier published alongside EchoPrime \cite{vukadinovic2024} was evaluated on each echo dataset. The log-softmax output of the model at the correct view (either apical four chamber or apical two chamber) was used to isolate the model's confidence as a function of image perturbation.

\paragraph{Cardiac Chamber Segmentation} UltraSam \cite{ultrasam} was used to segment the left ventricle and left atrium with increasing deformation. Since UltraSam is a segment anything model \cite{sam}, it relies on a bounding box prompt to segment images. This makes UltraSam far more robust to image degradations, and is more aligned with the strong human capacity for perception of images as they undergo degradation. Bounding boxes were extracted from geometrically warped true labels and provided to UltraSam (see Fig. S2 in supplementary).

\subsection{Image Reconstruction}
We trained implicit neural representations (INRs) for the USOVA 3D Follicles \cite{potocnik2020} and MicroSegNet Prostate \cite{jiang2024} datasets, using SIREN activations \cite{siren}, the Adam optimizer, and an initial learning rate of $10^{-3}$ annealed to $10^{-6}$ over the course of 300 epochs according to a cosine schedule. Each INR had a layer width of 256 with six intermediate SIREN layers, and skip connections at every even intermediate layer. A final tanh activation mapped the output to [-1, 1]. INRs were trained for each of the 16 follicle volumes and 75 prostate volumes using $L_2$ loss with an additional candidate LPIPS metric. We elected to exclude MedSAM from this analysis, as its input $1024^2$ resolution is impractical to use as a loss function for such tasks. After training, SSIM \cite{ssim} was calculated on each frame of the volume to measure the reconstruction quality of the perceptual metric. To emphasize the speckle accuracy, we also computed the high-frequency error norm (HFEN) as was done in \cite{adamson2025}, using $\sigma =2.5$.

\figSpearmanCorrelations

Although SSIM and HFEN are useful image reconstruction metrics for measuring the recovery of structure by a trained INR, they do not measure how realistic is a reconstructed image, which is especially notable in ultrasound reconstruction due to the unique distribution of speckles. To isolate these effects, we also computed the gray-level co-occurrence matrix (GLCM) \cite{glcm} to quantify texture realism. Five standard features were extracted from a symmetric GLCM computed at distances of 1 and 3 and angles $0^\circ$--$135^\circ$ with 32 gray levels: contrast, dissimilarity, homogeneity, energy, and correlation. GLCM features were used to measure the realism of reconstructed images by calculating the $L_1$ distance between the reference GLCM features and the reconstructed ones.

\subsection{Results}
\subsubsection{Supervised tasks}
\tabStats

\paragraph{Classification}
Spearman correlation and standard deviation of the various distance metrics to downstream classifier confidence is shown in Table S1 (supplementary) for the various deformations in our study, calculated across the axis of degradation per-frame and averaged across datasets. The three CNN models in our dataset (AlexNet, VGG-16, RadImageNet) all had high correlation to downstream confidence, outperforming classical metrics and several of the ViT models. However, the highest correlations were achieved by ultrasound models, which were more correlated to downstream confidence than other ViT variants.
Among CLIP models, BiomedCLIP and UltrasoundCLIP both outperformed CLIP, suggesting that exposure to ultrasound images may improve correlation. Surprisingly, MedSAM and USFM, both ViT-b variants, were among the least correlated models and were even outperformed by the ImageNet ViT.

\paragraph{Segmentation}
Spearman correlation and standard deviation of the various distance metrics to downstream dice score is shown in Table S2 (supplementary) for the various deformations in our study, calculated across the axis of degradation per-frame and averaged across datasets. Interestingly, all models saw marginal but clear improvement in correlation to rotation relative to classification, but failed entirely to correlate when tested on depth deformations. This is because both CAMUS and EchoNet are within UltraSam's training data, and small perturbation of depth creates downsized heart chambers that are still within the distribution of the dataset. Other augmentations like rotation and sector width immediately caused significant drops in UltraSam's dice score leading to higher correlation with perceptual metrics.

\figInrFollicles

Major trends noted in classification remained the same in segmentation, with TUSA and Ultrasound-CLIP scoring the highest correlations followed by CNN models, and BiomedCLIP. The weak correlation of MedSAM to segmentation is particularly notable because it is a segmentation model and was fine-tuned from the same source model as UltraSam. The failure of MedSAM to correlate with the performance of such a similar model may indicate that its training data is heavily biased towards other radiology applications, or that the image embedding alone does not contain any indication of uncertainty, which may be induced by cross-attention with prompts provided to SAM models.

Overall distance metrics and standard deviations averaged across the various augmentations are shown for each model on each supervised task and dataset in fig.~1. CNNs or models with high exposure to ultrasound data tended to outperform supervised alternatives, with the exception of MedSAM and USFM. Only a handful of models consistently beat $L_2$ loss, indicating that not all backbones are useful compared to classical metrics. SSIM was particularly noisy, as it sometimes correlated negatively with downstream metrics.

To provide a statistical analysis of the different backbone groups, we evaluate the null hypothesis that ultrasound foundation model metrics are no more correlated to downstream performance than each of the other metric groups using a Wilcoxon signed-rank test, and report the p-values in table~2. We reject the null hypothesis for classical metrics and LPIPS with natural images, but not for radiology models. One explanation for this is that although ultrasound is outnumbered in radiology datasets, radiology models still benefit from some degree of exposure to ultrasound data. Nonetheless, it is clear that LPIPS metrics do not uniformly correlate with downstream model perception, suggesting that backbone choice may be an important implementation choice.

\tabINRGrouped

\subsubsection{INR Reconstruction}
A sample image is shown from each distance metric (with the exclusion of MedSAM) in fig.~2. $L_2$ loss provides faithful image restoration but loses speckle quality, as models converge to the low-frequency solution. This effect is largely overcome by SSIM, which forces the INR to include high detail information. LPIPS from natural and medical generalist backbones induces hallucination of fine details, bending the tissue between the two largest follicles, distorting acoustic shadow, or failing to reconstruct the image altogether. All of the ultrasound backbones faithfully reconstruct input image with various levels of detail. USFM and Ultrasound-CLIP generate highly realistic speckle detail reminiscent of the SSIM reconstruction baseline.

SSIM, HFEN, and GLCM scores for each INR family are shown in table~3. $L_2$ loss produces the best reconstruction (SSIM and HFEN), but blurs the ultrasound texture. Medical backbones score lower in reconstruction but higher in realism. This effect is visible in fig.~2 where models like BiomedCLIP and RadImageNet that fail to reconstruct the image still create meaningful texture. Ultrasound backbones provide a balance between reconstruction quality and realistic texture, suggesting that they may be strong candidates for LPIPS backbones in reconstruction tasks.

\section{Conclusion}

In this paper, we studied the importance of ultrasound models in perceptual metrics for B-mode ultrasound images. We found that in supervised tasks, CNN backbones were highly correlated with downstream performance, but the highest correlations were achieved by ViT variants with exposure to ultrasound images. This was especially notable in CLIP models that were all trained in the same framework with the same architecture, but had increasing improvement in both downstream correlation and reconstruction with additional exposure to ultrasound images. In image reconstruction, ultrasound models provided a balance between reconstruction quality and realistic texture, and were the only LPIPS backbones we tested that did not distort anatomic details upon qualitative inspection. This work is a first step in challenging the widespread usage of arbitrary LPIPS backbones, but it is important to note that correlation to model performance is a proxy for visual perception, and further work is needed to quantify the relationship between human perception of B-mode images and the various LPIPS backbones studied here. Additional work can be done on adding more datasets and downstream models to be tested, but this is currently limited by the need for realistic augmentations, which do not meaningfully exist outside of echocardiography. Nonetheless, our findings suggest that exposure to ultrasound images may improve the suitability of models to serve as LPIPS backbones.

\bibliographystyle{splncs04}
\bibliography{main}

@String(CVPR  = {IEEE Conf. Comput. Vis. Pattern Recog.})

@String(ICCV  = {Int. Conf. Comput. Vis.})

@String(NeurIPS = {Adv. Neural Inform. Process. Syst.})

@String(CVPR  = {CVPR})

@String(ICCV  = {ICCV})

@String(NeurIPS = {NeurIPS})

@article{glcm,
author  = {Haralick, Robert M. and Shanmugam, Karthikeyan and Dinstein, Itzhak},
title   = {Textural Features for Image Classification},
journal = {IEEE Transactions on Systems, Man, and Cybernetics},
year    = {1973},
volume  = {SMC-3},
number  = {6},
pages   = {610--621},
doi     = {10.1109/TSMC.1973.4309314}
}

@article{camus,
author  = {Leclerc, S. and Smistad, E. and Pedrosa, J. and Ostvik, A. and Cervenansky, F. and Espinosa, F. and Espeland, T. and Berg, E. A. R. and Jodoin, P. and Grenier, T. and L{\o}vstakken, L. and Bernard, O.},
title   = {Deep Learning for Segmentation Using an Open Large-Scale Dataset in 2D Echocardiography},
journal = {IEEE Transactions on Medical Imaging},
year    = {2019},
volume  = {38},
number  = {9},
pages   = {2198--2210},
doi     = {10.1109/TMI.2019.2900516}
}

@article{echonet,
author  = {Ouyang, David and He, Bryan and Ghorbani, Amirata and Lungren, Matthew P. and Ashley, Euan A. and Liang, David H. and Zou, James Y.},
title   = {Video-based AI for beat-to-beat assessment of cardiac function},
journal = {Nature},
year    = {2020},
volume  = {580},
number  = {7802},
pages   = {252--256},
doi     = {10.1038/s41586-020-2145-8}
}

@misc{echogains,
author    = {Van De Vyver, Gilles and Lenz, Aksel Try and Smistad, Erik and Olaisen, Sindre Hellum and Grenne, Bjørnar and Holte, Espen and Dalen, Håvard and Løvstakken, Lasse},
title     = {Generative augmentations for improved cardiac ultrasound segmentation using diffusion models},
publisher = {arXiv},
year      = {2025},
month     = {Feb},
note      = {arXiv:2502.20100},
url       = {https://arxiv.org/abs/2502.20100}
}

@inproceedings{clip,
author    = {Radford, Alec and Kim, Jong Wook and Hallacy, Chris and Ramesh, Aditya and Goh, Gabriel and Agarwal, Sandhini and Sastry, Girish and Askell, Amanda and Mishkin, Pamela and Clark, Jack and Krueger, Gretchen and Sutskever, Ilya},
title     = {Learning Transferable Visual Models From Natural Language Supervision},
booktitle = {Proceedings of the 38th International Conference on Machine Learning},
series    = {Proceedings of Machine Learning Research},
volume    = {139},
pages     = {8748--8763},
year      = {2021},
publisher = {PMLR},
url       = {https://proceedings.mlr.press/v139/radford21a.html}
}

@inproceedings{ultrasound-clip,
author    = {Jin, Jiayun and Chai, Haolong and Huang, Xueying and Guo, Xiaoqing and Zheng, Zengwei and Zhou, Zhan and Wang, Junmei and Wang, Xinyu and Liu, Jie and Zhou, Binbin},
title     = {Ultrasound-CLIP: Semantic-Aware Contrastive Pre-training for Ultrasound Image-Text Understanding},
booktitle = CVPR,
year      = {2026},
doi       = {10.48550/arXiv.2604.01749},
url       = {https://arxiv.org/abs/2604.01749}
}

@misc{biomedCLIP,
author    = {Zhang, Sheng and Xu, Yanbo and Usuyama, Naoto and Xu, Hanwen and Bagga, Jaspreet and Tinn, Robert and Preston, Sam and Rao, Rajesh and Wei, Mu and Valluri, Naveen and Wong, Cliff and Tupini, Andrea and Wang, Yu and Mazzola, Matt and Shukla, Swadheen and Liden, Lars and Gao, Jianfeng and Crabtree, Angela and Piening, Brian and Bifulco, Carlo and Lungren, Matthew P. and Naumann, Tristan and Wang, Sheng and Poon, Hoifung},
title     = {BiomedCLIP: a multimodal biomedical foundation model pretrained from fifteen million scientific image-text pairs},
publisher = {arXiv},
year      = {2023},
doi       = {10.48550/arXiv.2303.00915},
url       = {https://arxiv.org/abs/2303.00915}
}

@inproceedings{lpips,
title={The Unreasonable Effectiveness of Deep Features as a Perceptual Metric},
author={Zhang, Richard and Isola, Phillip and Efros, Alexei A. and Shechtman, Eli and Wang, Oliver},
booktitle = CVPR,
year={2018}
}

@inproceedings{ldm,
title={High-Resolution Image Synthesis with Latent Diffusion Models},
author={Rombach, Robin and Blattmann, Andreas and Lorenz, Dominik and Esser, Patrick and Ommer, Bjorn},
booktitle = CVPR,
year={2022}
}

@article{grutman2025,
title={Implicit neural representation for scalable 3D reconstruction from sparse ultrasound images},
author={Grutman, Tal and Bismuth, Mike and Glickstein, Bar and Ilovitsh, Tali},
journal={npj Acoustics},
volume={1},
number={14},
pages={1--11},
year={2025},
publisher={Nature Publishing Group}
}

@misc{grutman2026,
    title={A texture-based framework for foundational ultrasound models}, 
    author={Tal Grutman and Carmel Shinar and Tali Ilovitsh},
    year={2026},
    eprint={2602.01444},
    archivePrefix={arXiv},
    primaryClass={eess.IV},
    url={https://arxiv.org/abs/2602.01444}, 
}

@article{radimagenet,
title={RadImageNet: An Open Radiologic Deep Learning Research Dataset for Effective Transfer Learning},
author={Mei, Xueyan and Liu, Zelong and Robson, Philip M. and Marinelli, Bradley and Huang, Mingqian and Doshi, Amish and Jacobi, Adam and Cao, Cheng and Link, Thomas M. and Yang, Yi and others},
journal={Radiology: Artificial Intelligence},
volume={4},
number={5},
pages={e210315},
year={2022},
publisher={Radiological Society of North America}
}

@article{cahan2025,
title={X-ray2CTPA: leveraging diffusion models to enhance pulmonary embolism classification},
author={Cahan, Noa and Klang, Eyal and Aviram, Galit and Barash, Yiftach and Konen, Eli and Giryes, Raja and Greenspan, Hayit},
journal={npj Digital Medicine},
volume={8},
number={439},
year={2025},
publisher={Nature Publishing Group}
}

@article{adamson2025,
title={Using deep feature distances for evaluating the perceptual quality of {MR} image reconstructions},
author={Adamson, Philip M and Desai, Arjun D and Dominic, Jeffrey and Varma, Maya and Bluethgen, Christian and Wood, Jeff P and Syed, Ali B and Boutin, Robert D and Stevens, Kathryn J and Vasanawala, Shreyas and Pauly, John M and Gunel, Beliz and Chaudhari, Akshay S},
journal={Magnetic Resonance in Medicine},
volume={94},
number={1},
pages={317--330},
year={2025},
publisher={Wiley Online Library}
}

@inproceedings{transfusion,
title={Transfusion: Understanding Transfer Learning for Medical Imaging},
author={Raghu, Maithra and Zhang, Chiyuan and Kleinberg, Jon and Bengio, Samy},
booktitle = NeurIPS,
year={2019}
}

@article{yuan2025,
title={Rethinking Domain-Specific Pretraining by Supervised or Self-Supervised Learning for Chest Radiograph Classification: A Comparative Study Against {ImageNet} Counterparts in Cold-Start Active Learning},
author={Yuan, Han and Zhu, Mingcheng and Yang, Rui and Liu, Han and Li, Irene and Hong, Chuan},
journal={Health Care Science},
volume={4},
number={2},
pages={110--143},
year={2025},
publisher={Wiley Online Library}
}

@article{zu2025,
title={Pre-trained Models Succeed in Medical Imaging with Representation Similarity Degradation},
author={Zhang, J. and Smith, L. and others},
journal={arXiv preprint arXiv:2503.07958},
year={2025}
}

@article{ultrasam,
title={{UltraSAM}: a foundation model for ultrasound using large open-access segmentation datasets},
author={Meyer, Adrien and Murali, Aditya and Zarin, Farahdiba and Mutter, Didier and Padoy, Nicolas},
journal={International Journal of Computer Assisted Radiology and Surgery},
volume={20},
number={1},
pages={1--10},
year={2025},
publisher={Springer}
}

@article{medsam,
title={Segment anything in medical images},
author={Ma, Jun and He, Yuting and Li, Feifei and Han, Lin and You, Chenyu and Wang, Bo},
journal={Nature Communications},
volume={15},
number={1},
pages={654},
year={2024},
publisher={Nature Publishing Group}
}

@article{usfm,
title={{USFM}: A universal ultrasound foundation model generalized to tasks and organs towards label efficient image analysis},
author={Jiao, Jing and Zhou, Jin and Li, Xiaokang and Xia, Menghua and Huang, Yi and Huang, Lihong and Wang, Na and Zhang, Xiaofan and Zhou, Shichong and Wang, Yuanyuan and Guo, Yi},
journal={Medical Image Analysis},
volume={96},
pages={103202},
year={2024},
publisher={Elsevier}
}

@article{li2025,
title={Speckle2Self: Self-supervised ultrasound speckle reduction without clean data},
author={Li, Xuesong and Navab, Nassir and Jiang, Zhongliang},
journal={Medical Image Analysis},
pages={103755},
year={2025},
publisher={Elsevier},
}

@article{stevens2025,
title={High Volume Rate 3D Ultrasound Reconstruction with Diffusion Models},
author={Stevens, Tristan S. W. and Nolan, Oisín and Somphone, Oudom and Robert,
        Jean-Luc and van Sloun, Ruud J. G.},
year={2025},
journal={IEEE Transactions on Medical Imaging},
doi={10.1109/TMI.2025.3645849}
}

@article{wagner1983,
title={Statistics of speckle in ultrasound {B}-scans},
author={Wagner, Robert F and Smith, Stephen W and Sandke, John M and Lopez, Robert H},
journal={IEEE Transactions on Sonics and Ultrasonics},
volume={30},
number={3},
pages={156--163},
year={1983},
publisher={IEEE}
}

@article{ssim,
author={Zhou Wang and Bovik, A.C. and Sheikh, H.R. and Simoncelli, E.P.},
journal={IEEE Transactions on Image Processing}, 
title={Image quality assessment: from error visibility to structural similarity}, 
year={2004},
volume={13},
number={4},
pages={600-612},
doi={10.1109/TIP.2003.819861}
}

@inproceedings{swinvit,
title={Swin Transformer: Hierarchical Vision Transformer using Shifted Windows},
author={Liu, Ze and Lin, Yutong and Cao, Yue and Hu, Han and Wei, Yixuan and Zhang, Zheng and Lin, Stephen and Guo, Baining},
booktitle=ICCV,
year={2021}
}

@misc{vukadinovic2024,
    title={EchoPrime: A Multi-Video View-Informed Vision-Language Model for Comprehensive Echocardiography Interpretation}, 
    author={Milos Vukadinovic and Xiu Tang and Neal Yuan and Paul Cheng and Debiao Li and Susan Cheng and Bryan He and David Ouyang},
    year={2024},
    eprint={2410.09704},
    archivePrefix={arXiv},
    primaryClass={cs.CV},
    url={https://arxiv.org/abs/2410.09704}, 
}

@article{potocnik2020,
title={Public database for validation of follicle detection algorithms on 3D ultrasound images of ovaries},
author={Poto{\v{c}}nik, Bo{\v{z}}idar and Munda, Jurij and Relji{\v{c}}, Milan and Raki{\'c}, Ksenija and Knez, Jure and Vlaisavljevi{\'c}, Veljko and Sedej, Ga{\v{s}}per and Cigale, Boris and Holobar, Ale{\v{s}} and Zazula, Damjan},
journal={Computer Methods and Programs in Biomedicine},
volume={196},
pages={105621},
year={2020},
publisher={Elsevier}
}

@article{jiang2024,
title={{MicroSegNet}: A deep learning approach for prostate segmentation on micro-ultrasound images},
author={Jiang, Hongxu and Imran, Muhammad and Muralidharan, Preethika and Patel, Anjali and Pensa, Jake and Liang, Muxuan and Benidir, Tarik and Grajo, Joseph R. and Joseph, Jason P. and Terry, Russell and others},
journal={Computerized Medical Imaging and Graphics},
volume={112},
pages={102326},
year={2024},
publisher={Elsevier}
}

@misc{siren,
    title={Implicit Neural Representations with Periodic Activation Functions}, 
    author={Vincent Sitzmann and Julien N. P. Martel and Alexander W. Bergman and David B. Lindell and Gordon Wetzstein},
    year={2020},
    eprint={2006.09661},
    archivePrefix={arXiv},
    primaryClass={cs.CV},
    url={https://arxiv.org/abs/2006.09661}, 
}

@misc{sam,
    title={Segment Anything}, 
    author={Alexander Kirillov and Eric Mintun and Nikhila Ravi and Hanzi Mao and Chloe Rolland and Laura Gustafson and Tete Xiao and Spencer Whitehead and Alexander C. Berg and Wan-Yen Lo and Piotr Dollár and Ross Girshick},
    year={2023},
    eprint={2304.02643},
    archivePrefix={arXiv},
    primaryClass={cs.CV},
    url={https://arxiv.org/abs/2304.02643}, 
}

% ---- Supplementary Material ----
\clearpage
\section*{Supplementary Material}
\addcontentsline{toc}{section}{Supplementary Material}

% Prefix Figure/Table numbers with "S"
\setcounter{figure}{0}
\setcounter{table}{0}
\renewcommand{\thefigure}{S\arabic{figure}}
\renewcommand{\thetable}{S\arabic{table}}
\renewcommand{\theHfigure}{S\arabic{figure}}
\renewcommand{\theHtable}{S\arabic{table}}

\subsection*{Supplementary Figures}
\figEchoGains

\clearpage
\figUltrasamPredictions

\clearpage
\subsection*{Supplementary Tables}
\tabClassificationDetailed

\tabSegmentationDetailed

\end{document}